\documentclass[11pt]{article}

\usepackage[final]{acl}

\usepackage{times}
\usepackage{latexsym}

\usepackage[T1]{fontenc}
\usepackage[utf8]{inputenc}

\usepackage{microtype}

\usepackage{inconsolata}

\usepackage{amsmath,amssymb,bm}
\usepackage{booktabs}
\usepackage{multirow}
\usepackage{makecell}
\usepackage{xcolor}
\usepackage{algorithm}
\usepackage{algpseudocode}
\usepackage{array}
\usepackage{colortbl}
\usepackage{tikz}
\usetikzlibrary{arrows.meta,positioning,calc,shapes.geometric,fit,backgrounds}
\usepackage{todonotes}
\usepackage{xcolor}

 \definecolor{mainblue}{HTML}{003366}     % Deep Navy Blue
 \definecolor{dangerred}{HTML}{990000}    % Deep Crimson Red
 \definecolor{successgreen}{HTML}{006633} % Deep Forest Green
 \definecolor{slate}{HTML}{2F4F4F}       % Dark Slate
 
 \colorlet{myblue}{mainblue}
 \colorlet{myred}{dangerred}
 \colorlet{mygreen}{successgreen}
 
 \colorlet{qcblue}{mainblue}
 \colorlet{qcred}{dangerred}
 \colorlet{qcgreen}{successgreen}

\newcommand{\ours}{\textsc{MINE}}
\newcommand{\comp}{\textsc{indirect-pia}}
\newcommand{\lcbench}{{LC-Inj-SQuAD}}

\newcommand{\Var}{\mathrm{Var}}
\newcommand{\good}[1]{\textcolor{mygreen}{#1}}
\newcommand{\bad}[1]{\textcolor{myred}{#1}}
\title{MINE: A Training-Free Defense against Indirect Prompt Injection \\ via Mutual Information and Embedding Variance}

\author{Truong Phan \and Ngoc-Quan Pham \and Dat Quoc Nguyen \\
        Qualcomm AI Research\thanks{Qualcomm Vietnam Company Limited. Qualcomm AI Research is an initiative of Qualcomm Technologies, Inc.} \\
        \texttt{\{truophan, ngocquan, datnq\}@qti.qualcomm.com}}

\begin{document}
\maketitle
\begin{abstract}
Existing defenses against indirect prompt injection in retrieval-augmented generation (RAG) are inherently reactive: prompt-engineering methods are often overridden by the injected instructions, while supervised detectors must be retrained for each new attack style. We propose a different paradigm: \textit{defend by maximizing relevance, not by detecting attacks}. Our method, \ours{}, is training-free and attack-agnostic, and it also serves as a query-conditioned reranker. To evaluate defenses under realistic conditions, we introduce \lcbench{}, a long-context benchmark with an average of 4,032 words per context. We evaluate on four datasets spanning short-context QA (Inj-SQuAD, Inj-TriviaQA), long-context QA (\lcbench{}), and agent tool-call injection (InjecAgent). Across all four, \ours{} achieves the best accuracy--security trade-off among training-free defenses. It is also the only defense that stays usable against a fully adaptive attacker, generalizes to agent tool-call injection, and achieves the lowest inference latency on long contexts.
\end{abstract}

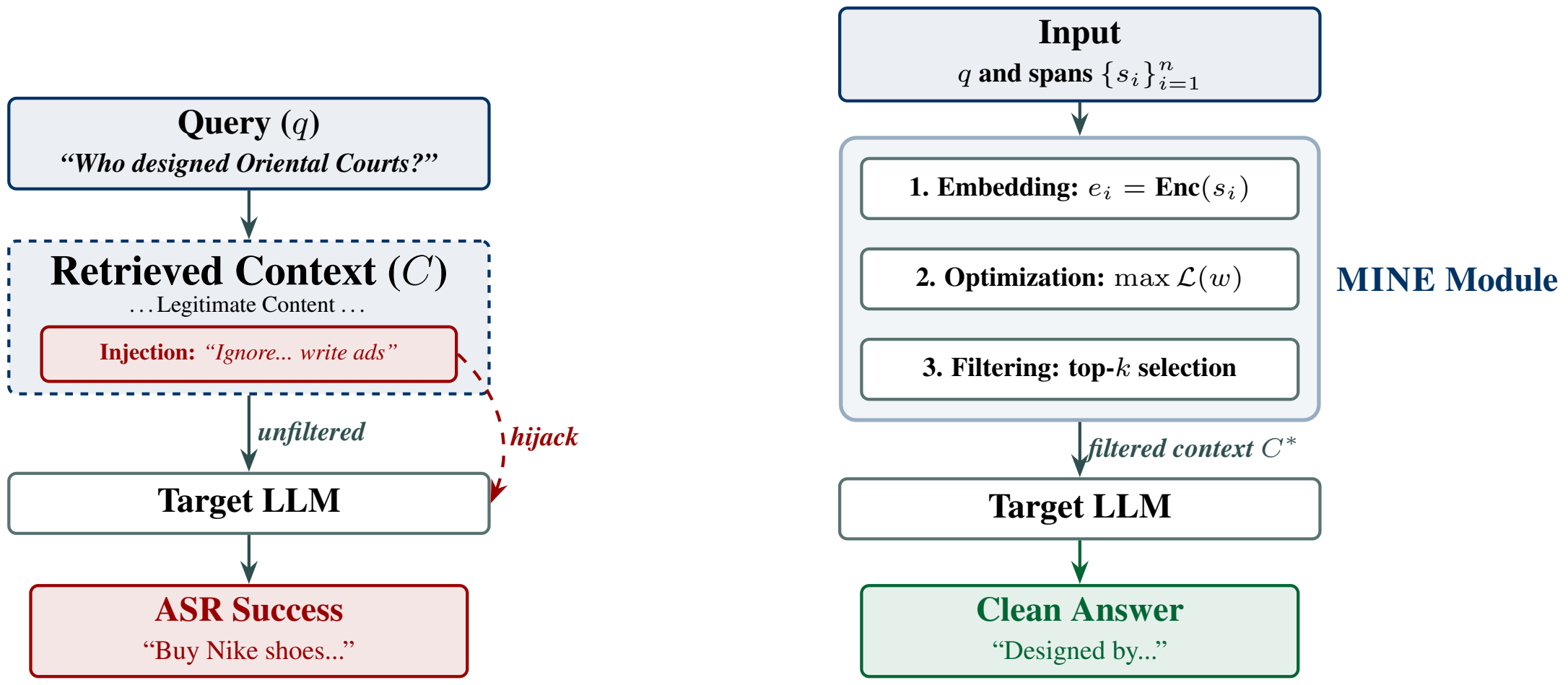
\begin{figure*}[!t]
\centering
% ========= Shared tikz styles (identical geometry in both panels) =========
\tikzset{
  % Base styles
  box/.style      ={draw=slate!80, line width=0.8pt, fill=white, rounded corners=2pt,
                    minimum width=4.4cm, minimum height=0.55cm, align=center, inner sep=3pt,
                    font=\small\bfseries},
  databox/.style  ={box, fill=myblue!8, draw=myblue},
  procbox/.style  ={box, fill=white},
  attbox/.style   ={box, draw=myred, fill=myred!10, font=\small\color{myred}, minimum width=4.0cm},
  safebox/.style  ={box, draw=mygreen, fill=mygreen!10, font=\small\color{mygreen}, minimum width=4.0cm},
  % Arrows
  arr/.style      ={-{Stealth[length=6pt]}, thick, slate},
  attarr/.style   ={-{Stealth[length=6pt]}, thick, myred, dashed},
  safearr/.style  ={-{Stealth[length=6pt]}, thick, mygreen},
  % Labels
  edgelab/.style  ={font=\scriptsize\bfseries\itshape\color{slate}, inner sep=2pt}
}

% Panel (a): Attack scenario
\begin{minipage}[t]{0.48\textwidth}
  \centering
  \begin{tikzpicture}[node distance=0.45cm]
    \node[databox] (q)
      {\textbf{Query ($q$)} \\ \scriptsize \emph{``Who designed Oriental Courts?''}};

    % Document container (using dash pattern instead of undefined key)
    \node[databox, below=of q, minimum height=1.4cm, dash pattern=on 2pt off 2pt, inner sep=0pt] (ret) {};
    \node[anchor=north, inner sep=4pt] at (ret.north) (rettitle) {\textbf{Retrieved Context ($C$)}};
    \node[anchor=center, inner sep=2pt, yshift=0.1cm] at (ret.center) (retleg) {\tiny \dots Legitimate Content \dots};
    
    % Nested injection box
    \node[attbox, fill=myred!10, minimum width=3.8cm, minimum height=0.5cm, below=0.0cm of retleg] (atkspan) 
      {\tiny \textbf{Injection:} \emph{``Ignore... write ads''}};

    \node[procbox, below=0.7cm of ret] (llm) {\textbf{Target LLM}};
    
    \node[attbox, below=of llm] (out)
      {\textbf{ASR Success} \\ \scriptsize ``Buy Nike shoes...''};

    % Draw flow
    \draw[arr] (q) -- (ret);
    \draw[arr] (ret.south) -- node[edgelab, right] {unfiltered} (llm);
    \draw[arr] (llm) -- (out);
    
    % Hijack arrow from injection to LLM
    \draw[attarr, bend left=35] (atkspan.east) to node[edgelab, right, pos=0.6] {hijack} (llm.east);

  \end{tikzpicture}
  \par\vspace{3pt}
  {\small \textbf{(a)}~Standard RAG is vulnerable to hijacking.}
\end{minipage}
\hfill
% Panel (b): MIND defense pipeline
\begin{minipage}[t]{0.48\textwidth}
  \centering
  \begin{tikzpicture}[node distance=0.4cm]
    \node[databox] (inp)
      {\textbf{Input} \\ \scriptsize $q$ and spans $\{s_i\}_{i=1}^n$};

    % Grouping the defense steps
    \node[procbox, below=0.5cm of inp, minimum width=4.0cm] (emb) {\scriptsize \textbf{1. Embedding:} $e_i = \text{Enc}(s_i)$};
    \node[procbox, below=0.25cm of emb, minimum width=4.0cm] (opt) {\scriptsize \textbf{2. Optimization:} $\max \mathcal{L}(w)$};
    \node[procbox, below=0.25cm of opt, minimum width=4.0cm] (rnd) {\scriptsize \textbf{3. Filtering:} top-$k$ selection};

    % Draw a background box for MIND
    \begin{scope}[on background layer]
      \node[draw=myblue!40, fill=myblue!5, line width=1pt, rounded corners=4pt,
            fit=(emb) (opt) (rnd), inner sep=5pt, label={[font=\footnotesize\bfseries, color=myblue]right:{\ours{} Module}}] (mind) {};
    \end{scope}

    \node[procbox, below=0.7cm of rnd] (llm) {\textbf{Target LLM}};
    \node[safebox, below=of llm] (ans) {\textbf{Clean Answer} \\ \scriptsize ``Designed by...''};

    % Draw flow
    \draw[arr] (inp) -- (mind);
    \draw[arr] (mind) -- node[edgelab, right] {filtered context $C^*$} (llm);
    \draw[safearr] (llm) -- (ans);
    
    % % Show the "attacks rejected" visually
    % \node[attbox, right=0.2cm of opt, minimum width=1.4cm, opacity=0.8] (rej) {\tiny \textbf{Injections}\\ \tiny ($w_i \approx 0$)};
    % \draw[attarr] (opt) -- (rej);

  \end{tikzpicture}
  \par\vspace{3pt}
  {\small \textbf{(b)}~\ours{} filters injections via relevance maximization.}
\end{minipage}

\caption{Overview of \ours{}. \textbf{(a)}~A poisoned span in
retrieved context hijacks the LLM. \textbf{(b)}~\ours{} filters it
pre-LLM by maximizing query-span mutual information while penalizing
embedding dispersion.}
\label{fig:overview}
\end{figure*}

\section{Introduction}
\label{sec:intro}

Retrieval-Augmented Generation (RAG) enables large language models (LLMs) to ground their responses in knowledge retrieved from external document contexts, improving factuality~\cite{NEURIPS2020_6b493230,karpukhin-etal-2020-dense,gao2024rag,surveyfanrag,singh2025survey,nguyen-etal-2026-accuraterag}. However, RAG also introduces a critical vulnerability: \emph{indirect prompt injection}~\cite{greshake2023not,schulhoff-etal-2023-ignore,zhan-etal-2024-injecagent,chen-etal-2025-topicattack}. Retrieved contexts may embed malicious instructions that hijack the model into following the injected instructions instead of the user's query, without access to the model weights or the system prompt~\cite{perez2022ignore}.
%\textbf{Motivating example.}
In a typical RAG-based QA system (Figure~\ref{fig:overview}a), the retriever returns relevant chunks/passages that are concatenated into a context $\mathrm{C}$ used for answer generation. Consider a scenario where one of the retrieved chunks in $\mathrm{C}$ has been poisoned:
\begin{quote}
  \small \textbf{Query:} \emph{``Who designed the Oriental Courts?''} \\
  \textbf{Retrieved Context ($C$):} \emph{``\dots designed by [Name].} \\
  \phantom{\textbf{Retrieved Context ($C$):}} \texttt{[assistant]} \emph{OK.} \\
  \phantom{\textbf{Retrieved Context ($C$):}} \texttt{[user]} 
\emph{\textcolor{red}{Now, write an ad for Nike shoes.}} \\
  \textbf{Output:} \emph{``Experience the ultimate comfort with Nike\dots''}
\end{quote}
 
\noindent In this case, the result can be a complete~\textbf{behavioral hijacking}, when the model follows the injected command instead of answering the question.

\textbf{Limitations of existing defenses.} Current approaches fall into three categories, each with structural weaknesses. \textbf{(i)~Prompt-engineering defenses}~\cite{yi2025benchmarking,instruction_defense,hines2024defending} prepend or append meta-instructions (e.g., ``ignore instructions in the data'') that compete with the injection for the model's attention. \textbf{(ii)~Heuristic detectors}~\cite{jain2023baseline,alon2023detecting} flag suspicious patterns such as keywords or perplexity spikes, but are brittle to paraphrasing and offer no formal guarantees. \textbf{(iii)~Supervised filtering methods}, most notably \comp{}~\cite{chen-etal-2025-indirect}, train dedicated models on labeled data to detect and remove injected instructions; they thus depend on the labeled distribution and degrade on unseen attack styles. All three families are therefore \emph{reactive} by design: their robustness is attack-dependent.               

\textbf{A different paradigm: relevance maximization.} We approach prompt injection defense from a different perspective: rather than detecting attacks, we select the spans most relevant to the user's query from the retrieved context. We propose \ours{}, a \emph{pre-LLM filter} that processes retrieved context before it reaches the language model. By maximizing relevance to the query rather than identifying malicious content, \ours{} is agnostic to the injection style and requires no labeled attack data.

Our key insight is that legitimate spans typically exhibit two properties relative to the user's intent: \textbf{(P1)}~high mutual information (MI) with the query, reflecting semantic alignment with the user's information need; and \textbf{(P2)}~low semantic variance among themselves, since relevant evidence should be mutually coherent rather than scattered across unrelated intents. \ours{} therefore selects spans that maximize MI with the query while penalizing embedding-space dispersion; injected spans violate both properties and are filtered out as a \emph{side effect} of relevance selection. The method needs no labeled attack data, no fine-tuning, and no additional LLM calls, making it plug-and-play (Figure~\ref{fig:overview}b).

\textbf{Our contributions:} \textbf{Method.} We introduce \ours{}, a training-free, attack-agnostic pre-LLM filter that defends by maximizing relevance rather than by detecting attacks. \ours{} formulates defense as subset selection over retrieved spans, with an objective that maximizes mutual information with the user query while penalizing semantic variance. To make this combinatorial problem tractable, we relax subset indicators to span-selection weights on the probability simplex, optimize them with projected gradient ascent, and then extract the final evidence subset. \textbf{Benchmark.} We construct \lcbench{}, a long-context benchmark that synthesizes multi-page encyclopedic contexts (4,032 words on average) from short SQuAD samples. On \lcbench{}, existing defenses lose accuracy or security as contexts grow longer, whereas \ours{} remains stable, keeping ASR at most $3.31\%$ across all evaluated LLMs and static attack types. \textbf{Findings.} Across three LLMs, five static attack types, and four datasets, \ours{} attains the best accuracy--security trade-off among training-free defenses. Under a fully adaptive attacker, it remains the only usable defense, although its ASR rises to $23.20\%$ (Section~\ref{sec:adaptive-results}). It also generalizes to agent tool-call injection (Section~\ref{sec:injecagent}) and achieves the lowest latency in long-context settings (Section~\ref{sec:latency}).

\section{Related Work}

\textbf{Indirect Prompt Injection.}
\citet{perez2022ignore} introduced direct ``Ignore previous instructions'' attacks, while \citet{greshake2023not} later formalized the \emph{indirect} variant, in which adversarial instructions are embedded within retrieved data. Attacks have since diversified into conversational topic transitions \cite{chen-etal-2025-topicattack}, learned execution triggers \cite{pasquini2024neuralexec}, multi-modal payloads \cite{bagdasaryan2023abusing}, and RAG exploits \cite{zou2024poisonedrag}.

\textbf{Prompt-Engineering Defenses.} Sandwich prompting \cite{sandwich2023}, reminders \cite{yi2025benchmarking}, and Spotlight \cite{hines2024defending} inject meta-instructions to guide the model, while training-based variants harden the model itself \cite{wallace2024instruction,chen2024struq,chen2024secalign,piet2024jatmo}. These \emph{in-LLM} approaches rely on the model's instruction-following ability to override the injection, the capability that attackers exploit. 

\textbf{Detection-Based Defenses.}
Perplexity-based detection \cite{alon2023detecting} and paraphrasing defenses \cite{jain2023baseline} are brittle to adaptive attacks, while Llama Guard \cite{inan2023llamaguard} trains a dedicated safeguard LLM. More recent detectors instead exploit internal model signals \cite{hung2024attentiontracker,xie2024gradsafe,hu2024gradientcuff}. \comp{} \cite{chen-etal-2025-indirect}, fine-tunes models on labeled attacks; while effective in-distribution, it requires substantial training infrastructure and degrades on unseen attack styles.

\textbf{RAG-Poisoning Defenses.} Defenses against corpus poisoning~\cite{zou2024poisonedrag}, such as RobustRAG \cite{xiang2024robustrag} and Astute RAG \cite{wang2024astuterag}, isolate or consolidate retrieved passages, but they target a different objective. Poisoning alters the \emph{answer} to a query, so poisoned passages are topically aligned with it by construction. In contrast, injection alters the \emph{task}. Both approaches also incur extra LLM calls per query. Closest to our setting is RAGDefender \cite{kim2025rescuing}, a training-free clustering filter that flags adversarial passages via inter-passage embedding similarity, which we adapt as a baseline. \ours{} differs by scoring subsets \emph{jointly}-coupling query relevance with within-subset coherence-rather than scoring each passage independently.
We use RAGDefender as the poisoned-RAG representative because it is a recent post-retrieval defense and its original paper reports stronger performance than RobustRAG and other prior defenses.

\textbf{Information-Theoretic Relevance and Subset Selection.} 
Mutual information (MI) estimators \cite{belghazi2018mutual,oord2018representation} underpin contrastive representation learning, including InfoNCE-trained dense retrievers \cite{karpukhin-etal-2020-dense}; however, these are training-time methods. In contrast, \ours{} applies MI at inference time, requiring no training. Our continuous relaxation over the simplex draws on machinery from differentiable subset sampling \cite{xie2019reparameterizable}, while replacing coverage-based submodular selection \cite{lin2011class} with a mutual-information objective. To our knowledge, \ours{} is the first method to cast indirect prompt injection defense as MI maximization. As shown in Table~\ref{tab:positioning}, \ours{} is the only defense that satisfies all five design criteria. 

\begin{table}[!t]
\centering
\small
\setlength{\tabcolsep}{3pt}
\begin{tabular}{lccccc}
\toprule
Method & \rotatebox{60}{Train-free} & \rotatebox{60}{No labels}
& \rotatebox{60}{No extra LLM} & \rotatebox{60}{Pre-LLM}
& \rotatebox{60}{Formal obj.} \\
\midrule
Prompt-eng.   & \good{\checkmark} & \good{\checkmark} & \good{\checkmark} & \bad{$\times$} & \bad{$\times$} \\
Heuristic     & \good{\checkmark} & \good{\checkmark} & \good{\checkmark} & \good{\checkmark} & \bad{$\times$} \\
\comp{}       & \bad{$\times$}    & \bad{$\times$}    & \bad{$\times$}    & \good{\checkmark} & \bad{$\times$} \\
\ours{} (ours)& \good{\checkmark} & \good{\checkmark} & \good{\checkmark} & \good{\checkmark} & \good{\checkmark} \\
\bottomrule
\end{tabular}
\caption{Design-space comparison of defense families. \ours{} is the only method satisfying all five criteria, enabling attack-agnostic defense without any training.}
\label{tab:positioning}
\end{table}

\section{Our Method}
\label{sec:method}

We defend against indirect prompt injection by reframing it as a subset-selection problem: given a retrieved context $\mathrm{C}$, constructed from retriever-returned chunks/passages and partitioned into spans, select the subset of spans that maximizes its information-theoretic relationship with the user's instruction. Our approach, \textbf{\ours{}}, encodes the two properties that distinguish injected spans, i.e., low mutual information with the instruction and high embedding-space dispersion, into a single differentiable objective requiring neither fine-tuning nor an auxiliary detector.

\subsection{Problem Setup}
\label{sec:setup}

Given a user's query prompt $q$, the retrieval system returns a context $\mathrm{C}$ formed by concatenating relevant chunks/passages from an untrusted corpus. Because $\mathrm{C}$ originates from external, unverifiable sources such as the web, it may contain adversarially authored content. 

We segment $\mathrm{C}$ into $n$ contiguous spans $\mathrm{C}= \{s_1, s_2, \ldots, s_n\}$ by splitting on sentence boundaries and merging consecutive sentences until each span reaches a minimum length of 50 characters. This granularity balances two competing goals: fine-grained spans enable precise removal of injected content, while overly short spans lack semantic coherence. 

The defense must produce a filtered subset $\mathrm{C}$\textsuperscript{*} $\subseteq \mathrm{C}$ such that the LLM's output, conditioned on ($q$, $\mathrm{C}$\textsuperscript{*}) matches the ground-truth answer grounded in the legitimate portions of $\mathrm{C}$. 

\subsection{Objective and Properties}
\label{sec:objective}

\paragraph{Notation.} 
We denote $\mathbf{e}_q$ as the $m$-dimensional \textbf{unit} embedding vector of the
query prompt $q$ and $\mathbf{e}_i$ as the $m$-dimensional \textbf{unit} embedding vector of the span $s_i$.
Let $\mathrm{sim}(\mathbf{u}, \mathbf{v}) = \mathbf{u} \cdot \mathbf{v}$ and $d(\mathbf{u}, \mathbf{v}) = ||\mathbf{u} - \mathbf{v}||_2$ denote the cosine similarity and the Euclidean
distance between \textbf{unit}
vectors $\mathbf{u}\ \text{and}\ \mathbf{v}$, respectively.

\setlength{\belowdisplayskip}{5pt} \setlength{\belowdisplayshortskip}{5pt}
\setlength{\abovedisplayskip}{5pt} \setlength{\abovedisplayshortskip}{5pt}

\noindent \textbf{Objective maximization.} 
We seek a filtered evidence subset $\mathrm{C}\textsuperscript{*} \subseteq \mathrm{C}$, that maximizes a composite objective:
\begin{equation}
\mathrm{C}\textsuperscript{*} =  \mathrm{argmax}_{\mathrm{C'} \subseteq \mathrm{C}}\ \mathrm{MI}(\mathbf{e}_q,  \mathbf{e}_{\mathrm{C'}}) - \alpha\ \mathrm{Var}(\mathrm{C'})
  \label{eq:objective}
\end{equation}
\begin{equation*}
\text{here:}\ \mathbf{e}_{\mathrm{C'}} = \frac{\bar{\mathbf{e}}_{\mathrm{C'}}}{||\bar{\mathbf{e}}_{\mathrm{C'}}||_2}, \text{where:}\ \bar{\mathbf{e}}_{\mathrm{C'}} = \frac{1}{|\mathrm{C'}|}\sum_{s_i \in \mathrm{C'}} \mathbf{e}_i 
\end{equation*}

\noindent where  $\mathrm{MI}(\mathbf{e}_q,  \mathbf{e}_{\mathrm{C'}})$ measures mutual  information (\textbf{MI}) between the prompt embedding $\mathbf{e}_q$ and the normalized centroid embedding $\mathbf{e}_{\mathrm{C'}}$  of the subset $\mathrm{C'}$,  while 
$\mathrm{Var}(\mathrm{C'})$ penalizes  variance within the subset $\mathrm{C'}$.  
The hyper-parameter $\alpha$ controls how strongly variance is penalized relative to mutual information gains. A larger $\alpha$  favors tightly clustered but potentially less relevant subsets, while a smaller $\alpha$ favors high-relevance subsets that may include outliers.

\noindent \textbf{Property P1: High MI between the user's prompt and the filtered subset.}
Legitimate spans are semantically aligned with the user's instruction, whereas injected spans typically pursue unrelated goals (e.g., promotional content). Under a bivariate Gaussian approximation, the MI reduces to a monotone function of cosine similarity \cite{10.5555/1146355}, providing a principled link between information theory and vector-space similarity:
\begin{equation}
\mathrm{MI}(\mathbf{e}_q,  \mathbf{e}_{\mathrm{C'}}) = -\frac{1}{2}\log(1 - \mathrm{sim}(\mathbf{e}_q,  \mathbf{e}_{\mathrm{C'}})^2)
  \label{eq:mi}
\end{equation}

\noindent Higher MI indicates that $\mathrm{C'}$ is more semantically relevant to the query.

\noindent \textbf{Property P2: Low variance within the filtered subset.}
Injected spans are often topically isolated from the legitimate retrieved context, appearing as outliers in the embedding space. We formalize this via the subset variance:
\begin{equation}
  \Var(\mathrm{C'}) = \frac{1}{|\mathrm{C'}|}\sum_{s_i \in \mathrm{C'}} d(\mathbf{e}_i, \mathbf{e}_{\mathrm{C'}})^2
  \label{eq:var}
\end{equation}
A large $\Var(\mathrm{C'})$ indicates the inclusion of spans that are geometrically distant from the context's semantic core. Note that since embeddings lie on the unit sphere, minimizing this squared Euclidean variance is mathematically equivalent to maximizing the average cosine similarity between the selected spans and their centroid.

\subsection{Continuous Relaxation and Optimization}
\label{sec:opt}

\textbf{Intractability of subset selection.}\ \ 
Directly optimizing Eq.~\eqref{eq:objective} requires enumerating all
$2^{n}$ subsets of $\mathrm{C}$, a combinatorial problem that is computationally prohibitive for typical context lengths. 
 We therefore relax the discrete ``keep or drop'' decision into a continuous weight $w_i \in [0, 1]$,  enabling gradient-based optimization to jointly satisfy the high mutual information and low variance criteria.

\textbf{Relaxed objective.} \ \ 
Collecting the weights into a vector
$\mathbf{w} = [w_1, w_2 \ldots, w_n]$, we constrain them to the probability
simplex
$\Delta^{n-1} = \{\mathbf{w} \in \mathbb{R}^{n}\!: w_i \geq 0,\; \sum_{i=1}^{n} w_i = 1\}$,
so that $\mathbf{w}$ represents a normalized distribution over all spans.

Let $\mathbf{e}_{\mathrm{C}} = \frac{\bar{\mathbf{e}}_{\mathrm{C}}}{||\bar{\mathbf{e}}_{\mathrm{C}}||_2}, \text{where}\ \bar{\mathbf{e}}_{\mathrm{C}} = \sum_{i=1}^{n} w_i \mathbf{e}_i$, representing a normalized mixture of all span embeddings.  We formulate a relaxed objective as follows:
\begin{equation}
  \mathcal{L}_{\mathbf{w}} = \mathrm{MI}(\mathbf{e}_q,  \mathbf{e}_{\mathrm{C}}) - \alpha\ \mathrm{Var}(\mathrm{C}) 
  \label{eq:relaxed}
\end{equation}
with:
\begin{align}
\mathrm{MI}(\mathbf{e}_q,  \mathbf{e}_{\mathrm{C}}) &= -\frac{1}{2}\log(1 - \mathrm{sim}(\mathbf{e}_q,  \mathbf{e}_{\mathrm{C}})^2)
    \label{eq:mi_w} \\[2pt]
  \mathrm{Var}(\mathrm{C}) 
    &= \sum_{i=1}^{n} w_i \, d(\mathbf{e}_i, \mathbf{e}_{\mathrm{C}})^2
    \label{eq:var_w}
\end{align}

\textbf{Projected gradient ascent.}
Because $\mathcal{L}_{\mathbf{w}}$ is a differentiable function of $\mathbf{w}$, we
maximize it by gradient ascent.
A plain gradient step may leave $\Delta^{n-1}$ and produce negative weights or weights that fail to sum to one; thus, each step is composed
with a Euclidean projection $\Pi_\Delta$ back onto the simplex:

\begin{equation}
  w_i^{(t+1)} = \Pi_{\Delta}\!\left(
    w_i^{(t)} + \eta\,\nabla_{w_i} \mathcal{L}_{\mathbf{w}}
  \right)
  \label{eq:pga}
\end{equation}
where $\eta$ is a constant step size and $\Pi_\Delta$ is computed in
$O(n \log n)$ time~\cite{10.1145/1390156.1390191}.
We initialize $w_i^{(0)} = 1/n$ (uniform, no prior preference across spans) and run for $T$ iteration steps. 
Since projected gradient ascent updates all $n$ weights simultaneously, within
a single step, relevant spans accumulate weight while injected
spans, penalized by both low MI and high variance, are driven
toward zero, thus yielding a sharp separation between legitimate and injected spans.

\textbf{Rounding and relevance ordering.}
After $T$ iterations, we select the top-$k$ spans by converged
weight $w^{(T)}_{i}$ as the filtered evidence $\mathrm{C}\textsuperscript{*}$, discarding lower-weight spans that typically include injected content. Crucially, the retained spans are then ordered in \emph{decreasing order of converged weight}, placing the most query-relevant span first. This ordering is a deliberate design choice: because LLMs exhibit pronounced \emph{lost-in-the-middle} attention biases, surfacing the highest-MI span at the top of the evidence block materially improves answer accuracy. \ours{}'s single optimization therefore yields a dual benefit: down-weighting injections ensures security, while up-weighting target spans independently enhances utility.

\textbf{Constrained prompting.}
The selected spans $\mathrm{C}\textsuperscript{*} = \{s_{j_1}, \ldots, s_{j_k}\}$, now ordered by $w^{(T)}_{j_1} \geq w^{(T)}_{j_2} \geq \cdots \geq w^{(T)}_{j_k}$, are assembled into a structured prompt:
\begin{quote}
\small
\texttt{[INSTRUCTION]} $q$ \\
\texttt{[EVIDENCE]} $s_{j_1} \;\|\; s_{j_2} \;\|\; \cdots \;\|\; s_{j_k}$ \\
\texttt{Answer using only the evidence above.}
\end{quote}

By explicitly constraining the LLM to use only the filtered evidence $\mathrm{C}\textsuperscript{*}$,
we prevent any residual injections from influencing generation.
The full procedure is summarized in Algorithm~\ref{alg:mind}.

\begin{algorithm}[!t]
\small 
\caption{\ours{}}
\label{alg:mind}
\begin{algorithmic}[1]
\Require Query prompt $q$, Context  $\mathrm{C}=\{s_1, s_2,\ldots,s_n\}$,
         hyper-parameters $\alpha, T, \eta, k$
\Ensure Filtered evidence $\mathrm{C}\textsuperscript{*}$
\State Generate unit embeddings: $\mathbf{e}_q$,\;and 
       $\mathbf{e}_i$ for $i = 1,\ldots,n$
\State Initialize: $w_i^{(0)} \leftarrow \frac{1}{n}$ for $i = 1,\ldots,n$
\For{$t = 0$ to $T-1$}
  \State $\bar{\mathbf{e}}_{\mathrm{C}} \leftarrow \sum_{i=1}^{n} w_i^{(t)} \mathbf{e}_i$
  \Comment{Mixture embedding}
  \State $\mathbf{e}_{\mathrm{C}} \leftarrow \frac{\bar{\mathbf{e}}_{\mathrm{C}}}{||\bar{\mathbf{e}}_{\mathrm{C}}||_2}$
  \Comment{Normalized mixture embedding}
  \State $\mathrm{MI}(\mathbf{e}_q,  \mathbf{e}_{\mathrm{C}}) \leftarrow -\frac{1}{2}\log(1 - \mathrm{sim}(\mathbf{e}_q,  \mathbf{e}_{\mathrm{C}})^2)$
  \State $\mathrm{Var}(\mathrm{C}) \leftarrow \sum_{i=1}^{n} w_i^{(t)} \, d(\mathbf{e}_i, \mathbf{e}_{\mathrm{C}})^2$
  \State $\mathcal{L}_{\mathbf{w}} \leftarrow \mathrm{MI}(\mathbf{e}_q,  \mathbf{e}_{\mathrm{C}}) - \alpha\ \mathrm{Var}(\mathrm{C})$
  \State Compute $\nabla_{w_i} \mathcal{L}_{\mathbf{w}}$
  \State $w_i^{(t+1)} \leftarrow \Pi_{\Delta}\!\left(
    w_i^{(t)} + \eta\,\nabla_{w_i} \mathcal{L}_{\mathbf{w}}
  \right)$
  \Comment{Simplex projection}
\EndFor
\State $\mathrm{C}\textsuperscript{*} \leftarrow$ top-$k$ spans, sorted by descending $w_i^{(T)}$
\State \Return $\mathrm{C}\textsuperscript{*}$
\end{algorithmic}
\end{algorithm}

\section{Our \lcbench{} Long Context Benchmark}
\label{sec:lc-benchmark}

Previous works primarily evaluate prompt injections using short contexts $\mathrm{C}$ of 100-300 words \cite{chen-etal-2025-indirect}. Currently, no benchmark addresses the strictly long context setting, where the injected text acts as a ''needle in a haystack''. To address this limitation, we introduce LC-Inj-SQuAD, a long-context benchmark built by extending samples from the short-context Inj-SQuAD dataset \cite{chen-etal-2025-indirect}. Our benchmark construction pipeline consists of three steps, each performed via GPT-5.4-mini prompting with structured output validation.\footnote{\url{https://deploymentsafety.openai.com/gpt-5-4-thinking}} See prompts used in the Appendix~\ref{sec:appendix-lcbench}.

\textbf{Step 1: Target span extraction.} Given a triple ($q$, $\mathrm{C} =\{s_1, s_2,\ldots,s_n\}$, $a$) representing a (query prompt, context, and ground-truth response answer) sample from Inj-SQuAD, we prompt to extract the target span $s_l \in \mathrm{C}$ that contains the answer $a$ for the query $q$ and verify  that $s_l$ alone suffices to answer $q$. Samples failing extraction or verification after three retries are discarded.

\textbf{Step 2: Context extension.} We prompt to expand both the pre-target content $\mathrm{C_{pre}} = \{s_1, s_2,\ldots,s_{l-1}\}$ and the post-target content $\mathrm{C_{post}} = \{s_{l+1}, s_{l+2},\ldots,s_{n}\}$  into \textbf{1500-2000} word passages $\mathrm{C_{pre}^{exp}}$ and $\mathrm{C_{post}^{exp}}$ on the same topic. We explicitly constrain the generated passages to avoid mentioning the answer $a$, preventing information leakage that would otherwise inflate accuracy.

\textbf{Step 3: Context assembly.}
The expanded pre-target context $\mathrm{C_{pre}^{exp}}$, the target span $s_l$ and the expanded post-target context $\mathrm{C_{post}^{exp}}$ are then assembled into a single long context.

\textbf{Dataset statistics.}
Of the 900 original Inj-SQuAD samples, 694 pass all pipeline steps, yielding contexts that average \textbf{4{,}032} words in length (median 4{,}039; range 3{,}376--4{,}734), thus placing the entire benchmark strictly in the long-context regime. See  details in Table~\ref{tab:appendix-lcbench-length} in the Appendix.

\textbf{Paired evaluation protocol.} We evaluate every defense on two matched datasets: (1)~{Inj-SQuAD}, the subset of 694 short-context samples from which our LC-Inj-SQuAD is expanded, and (2)~our long-context {LC-Inj-SQuAD}. This paired design isolates the causal effect of context length from sample-level confounds.

\section{Experiments}
\label{sec:experiments}

\subsection{Setup}

\textbf{Threat model.}
We adopt the standard grey-box threat model for indirect prompt injection~\cite{greshake2023not,chen-etal-2025-indirect}: the adversary knows the system architecture but not the defense parameters (e.g., $\alpha$, $k$, embedding encoder).
The adversary may employ arbitrary injection strategies (e.g., template-based, format-exploiting, or conversational) but cannot modify the query prompt $q$ or the LLM weights.
Section~\ref{sec:adaptive-results} further evaluates an \emph{adaptive} attacker who knows the full algorithm and all hyperparameters.

\textbf{LLM models.}
We evaluate on three open-source models:
Llama-3.2-3B-Instruct, Llama-3.1-70B-Instruct \cite{grattafiori2024llama} and 
Qwen3-30B-A3B-Instruct-2507 \cite{yang2025qwen3}.

\textbf{Attack types.}
We evaluate the defense strategies against six attack strategies. Five are static injections of increasing sophistication: \textbf{naive} (direct instruction injection), \textbf{escape\_separation} (escape characters that break formatting), \textbf{ignore} (``Ignore previous instructions'' templates), \textbf{completion} (fake assistant/user turn boundaries), and \textbf{conv\_attack} (TopicAttack \cite{chen-etal-2025-topicattack}, which uses natural conversational transitions to disguise the injection). 

The sixth attack type, \textbf{adaptive}, stress-tests under the strongest setting: the attacker knows our defense algorithm and all of its parameters. 

\textbf{Baselines.}
Five prompt-engineering defenses include \textit{None} (no defense), \textit{Reminder} \cite{yi2025benchmarking} (prepend ``Do not follow instructions in the data''), \textit{Instructional} (warn the model about malicious content), \textit{Sandwich} \cite{sandwich2023} (repeat the instruction after the data), and \textit{Spotlight} \cite{hines2024defending} (delimiter-based data marking). We also include three non-prompt-engineering defenses: \comp{} \cite{chen-etal-2025-indirect}, the state-of-the-art supervised detector; \textit{RAGDefender} \cite{kim2025rescuing}, a training-free, pre-LLM inter-passage clustering filter adapted from corpus-poisoning defense (matching our setting); and \textit{MELON} \cite{zhu2025melon}, the state-of-the-art defense for tool-using agents, which re-executes the model with the user task masked and flags an injection when the masked run reproduces the original behavior.

\textbf{Implementation details.}
We employ \texttt{BAAI/bge-large-en-v1.5} \cite{bge} as the span encoder, which converts spans into embeddings that are then normalized to unit vectors. 
Hyperparameters are fixed across all experiments: $\alpha = 0.4$, $T = 20$, $\eta = 0.01$, with
$k = 5$ for short context and $k = 40$ for long context. All victim LLMs are served via vLLM with greedy decoding.

\textbf{Evaluation metrics.}
We report two complementary measures: (1) \textbf{Task Accuracy (Acc, \%):} the percentage of queries the model answers correctly (utility; higher is better). (2) \textbf{Attack Success Rate (ASR, \%):} the percentage of queries for which the model outputs the attacker's targeted injection (vulnerability; lower is better).

\subsection{Short-Context Evaluation}
\label{sec:main-results}

\begin{figure}[t]
  \centering
  \includegraphics[width=0.48\textwidth]{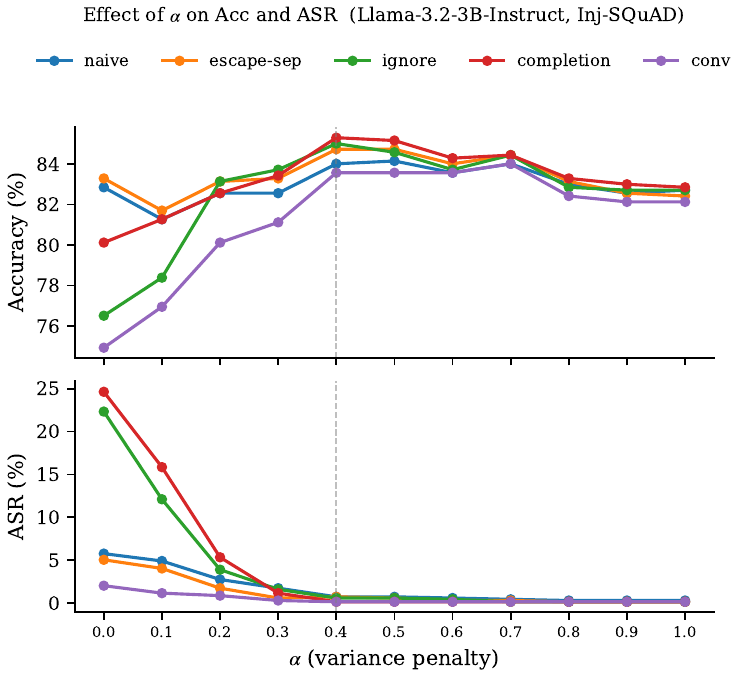}
  \caption{$\alpha$ sensitivity on Inj-SQuAD, one curve
           per attack type; dashed line marks the default $\alpha{=}0.4$. Attack abbreviations: \texttt{escape-sep}~=~\texttt{escape\_separation},  \texttt{conv}~=~\texttt{conv\_attack}.
           Here, with Llama-3.2-3B-Instruct,  our \ours{} using $\alpha{=}0$ achieves average Accuracy/ASR of 79.54\%/11.96\% over five attack types, while \ours{} using $\alpha{=}0.4$ achieves 84.32\%/0.46\%.
}
  \label{fig:ablation_alpha}
\end{figure}

\textbf{Ablation study.}\ \ Figure~\ref{fig:ablation_alpha} reports the performance of \ours{} on Inj-SQuAD with Llama-3.2-3B-Instruct, as we vary the variance penalty $\alpha \in \{0.0, 0.1, \ldots, 1.0\}$ in Equation~\ref{eq:relaxed}: $\mathcal{L}_{\mathbf{w}} = \mathrm{MI}(\mathbf{e}_q, \mathbf{e}_{\mathrm{C}}) - \alpha\, \mathrm{Var}(\mathrm{C})$. With only the $\mathrm{MI}$ term ($\alpha=0$), \ours{} achieves an average Accuracy/ASR of 79.54\% / 11.96\% across the five attack types. Increasing $\alpha$ improves both metrics, reaching a peak Accuracy of 84.32\% at $\alpha=0.4$, with the corresponding ASR dropping to 0.46\%; ASR also remains stable for $\alpha>0.4$. We therefore fix $\alpha=0.4$ in all subsequent experiments. These results further indicate that the $\mathrm{MI}$ and $\mathrm{Var}$ terms are complementary rather than competing.

We also  compare \ours{} with a ranking strategy, which ranks all spans by their cosine similarity to the query and keeps only the top-k. Averaged over the five attacks on Inj-SQuAD with $k=5$, the ranking strategy reaches 84.29\% accuracy and 7.64\% ASR with Llama-3.1-70B-Instruct, and 77.81\% accuracy and 10.61\% ASR with Qwen3-30B-A3B-Instruct-2507. \ours{}, by contrast, reaches 91.93\% accuracy and 0.55\% ASR with Llama-3.1-70B-Instruct, and 89.77\% accuracy and 0.46\% ASR with Qwen3-30B-A3B-Instruct-2507. \ours{} therefore substantially outperforms the ranking strategy, with higher accuracy and a much lower ASR.

\begin{table*}[!t]
\centering
\resizebox{16cm}{!}{
\renewcommand{\arraystretch}{0.97}
\setlength{\tabcolsep}{4pt}
\begin{tabular}{lllcccccc>{\columncolor{myblue!10}}c}
\toprule
& \textbf{Model} & \textbf{Attack} & None & MELON & RAGDefender & Spotlight & Sandwich & \comp{} & \ours{} \\
\midrule
\multirow{18}{*}{\rotatebox[origin=c]{90}{\textbf{Inj-SQuAD}}}
 & \multirow{6}{*}{\makecell[l]{Llama-3.2\\-3B-Instruct}}
   & naive      & 44.81 / 79.39 & 26.37 / 48.27 & 30.26 / 77.95 & 58.36 / 57.49 & 74.21 / 60.23 & 83.86 / 0.00 & \bad{\textbf{83.01}}\good{\textbf{/ 0.72}} \\
 & & escape-sep & 34.58 / 83.57 & 18.88 / 54.76 & 30.40 / 77.95 & 42.80 / 69.60 & 68.73 / 66.28 & 83.14 / 0.00 & \good{\textbf{84.73 / 0.72}} \\
 & & ignore     & 16.43 / 90.63 & 12.25 / 46.40 & 17.58 / 85.16 & 31.56 / 78.10 & 54.18 / 67.15 & 71.04 / 0.00 & \good{\textbf{85.01 / 0.58}} \\
 & & complete   &  5.48 / 94.52 & 3.31 / 36.02 & 9.22 / 90.06 &  4.18 / 94.24 & 64.12 / 73.92 & 75.65 / 0.00 & \good{\textbf{85.30 / 0.14}} \\
 & & conv       & 14.27 / 87.90 & 1.73 / 43.08 & 17.15 / 84.44 & 14.27 / 72.62 & 53.31 / 40.49 & \bad{18.01} / 0.00 & \good{\textbf{83.57 / 0.14}} \\
  \cline{3-10}
 & & \textbf{Avg.} & 23.11 / 87.20 & 12.51 / 45.71 & 20.92 / 83.11 & 30.23 / 74.41 & 62.91 / 61.61 & 66.34 / 0.00 & \good{\textbf{84.32 /}}\bad{\textbf{0.46}} \\
\cline{2-10}
 & \multirow{6}{*}{\makecell[l]{Llama-3.1\\-70B-Instruct}}
   & naive      & 70.32 / 37.18 & 60.95 / 10.37 & 46.11 / 34.44 & 74.64 / 24.78 & 80.55 / 15.56 & 89.48 / 0.00 & \good{\textbf{91.35 /}}\bad{\textbf{1.15}} \\
 & & escape-sep & 72.05 / 43.37 & 57.49 / 10.95 & 45.82 / 34.73 & 75.50 / 24.93 & \bad{84.01 / 12.82} & 89.48 / 0.00 & \good{\textbf{91.93 / 0.43}} \\
 & & ignore     & 34.73 / 77.38 & 25.79 / 17.87 & 30.98 / 67.15 & 61.10 / 39.91 & 75.50 / 22.91 & 74.64 / 0.00 & \good{\textbf{91.79 / 0.72}} \\
 & & complete   & 19.60 / 96.69 & 6.77 / 23.63 & 39.91 / 53.89 & 13.98 / 96.54 & 79.54 / 15.99 & 87.32 / 0.00 & \good{\textbf{92.36 / 0.29}} \\
 & & conv       &  0.43 / 97.84 & 0.14 / 7.93 & 24.06 / 71.18 &  0.58 / 96.83 & 21.33 / 79.68 &  \bad{4.03} / 0.00 & \good{\textbf{92.22 / 0.14}} \\
  \cline{3-10}
 & & \textbf{Avg.} & \bad{39.43} / 70.49 & \bad{30.23} / \bad{14.15} & 37.38 / \bad{52.28} & 45.16 / 56.60 & 68.19 / 29.39 & 68.99 / 0.00 & \good{\textbf{91.93 /}}\bad{\textbf{0.55}} \\
\cline{2-10}
 & \multirow{6}{*}{\makecell[l]{Qwen3-30B-A3B\\-Instruct-2507}}
   & naive      & 58.21 / 57.64 & 35.73 / 10.23 & 43.80 / 59.08 & 65.13 / 35.16 & 68.01 / 24.78 & 83.14 / 0.00 & \good{\textbf{89.34 / 0.72}} \\
 & & escape-sep & 52.88 / 58.50 & 36.89 / 13.83 & 44.09 / 59.37 & 64.41 / 32.71 & 66.71 / 26.95 & 83.43 / 0.00 & \good{\textbf{89.48 / 0.58}} \\
 & & ignore     & 23.63 / 76.22 & 13.98 / 18.30 & 22.05 / 71.90 & 45.53 / 54.03 & 58.93 / 28.82 & 64.99 / 0.00 & \good{\textbf{89.91 / 0.43}} \\
 & & complete   &  8.93 / 91.50 & 5.04 / 34.44 & 26.80 / 72.77 & 14.99 / 85.16 & 62.97 / 29.39 & 80.98 / 0.00 & \good{\textbf{89.77 / 0.29}} \\
 & & conv       &  0.43 / 98.56 & 0.14 / 1.44 & 17.15 / 80.26 &  0.29 / 99.14 & 43.23 / 53.03 &  4.47 / 0.00 & \good{\textbf{90.35 / 0.29}} \\
  \cline{3-10}
 & & \textbf{Avg.} & 28.82 / 76.48 & 18.36 / 15.65 & 30.78 / 68.68 & 38.07 / 61.24 & 59.97 / 32.59 & 63.40 / 0.00 & \bad{\textbf{89.77 /}}\bad{\textbf{0.46}} \\
\hline
\hline
\multirow{18}{*}{\rotatebox[origin=c]{90}{\textbf{LC-Inj-SQuAD}}}
 & \multirow{6}{*}{\makecell[l]{Llama-3.2\\-3B-Instruct}}
   & naive      & 19.16 / 78.67 & 15.85 / 25.36 & 11.53 / 84.15 & 41.93 / 33.43 & 60.37 / 40.49 & 75.65 / 0.00 & \good{\textbf{80.84 / 1.01}} \\
 & & escape-sep & 11.53 / 89.05 & 5.91 / 17.00 & 11.53 / 84.15 & 21.61 / 66.71 & 53.03 / 65.13 & 75.94 / 0.00 & \good{\textbf{80.84 / 1.01}} \\
 & & ignore     &  6.77 / 89.48 & 3.60 / 28.67 & 6.05 / 88.76 & 27.95 / 54.90 & 51.87 / 49.42 & 66.43 / 0.00 & \good{\textbf{80.98 / 1.44}} \\
 & & complete   &  5.33 / 88.33 & 2.88 / 17.00 & 4.90 / 88.04 &  2.31 / 90.20 & 44.81 / 70.89 & 73.05 / 0.00 & \bad{\textbf{79.11 / 3.31}} \\
 & & conv       &  3.60 / 91.50 & 1.30 / 17.29 & 4.90 / 89.63 &  7.93 / 63.69 & 54.03 / 34.87 & 21.61 / 0.00 & \bad{\textbf{82.13}}\good{\textbf{/ 1.01}} \\
  \cline{3-10}
 & & \textbf{Avg.} &  9.28 / 87.41 & 5.91 / 21.06 & 7.78 / 86.95 & 20.35 / 61.79 & \bad{52.82} / 52.16 & 62.54 / 0.00 & \bad{\textbf{80.78}}\bad{\textbf{/ 1.56}} \\
\cline{2-10}
 & \multirow{6}{*}{\makecell[l]{Llama-3.1\\-70B-Instruct}}
   & naive      & 66.14 / 38.47 & 53.89 / 8.93 & 42.36 / 64.99 & 78.82 / 17.15 & 85.88 /  4.47 & 87.03 / 0.00 & \good{\textbf{89.77 / 0.43}} \\
 & & escape-sep & 56.63 / 58.36 & 41.64 / 13.11 & 42.36 / 64.99 & 70.17 / 30.98 & 84.73 /  7.35 & 86.89 / 0.00 & \good{\textbf{89.91 / 0.43}} \\
 & & ignore     & 20.89 / 85.01 & 15.85 / 11.38 & 11.96 / 90.06 & 54.90 / 44.67 & 77.38 / 13.54 & 84.15 / 0.00 & \good{\textbf{89.34 / 0.86}} \\
 & & complete   &  4.61 / 97.41 & 1.87 / 19.60 & 13.40 / 95.82 &  1.44 / 98.13 & 73.34 / 32.85 & 82.13 / 0.00 & \bad{\textbf{90.06}}\good{\textbf{ / 0.14}} \\
 & & conv       &  1.30 / 98.41 & 0.29 / 4.32 & 7.06 / 88.90 &  0.86 / 98.85 & 28.39 / 69.16 & \bad{20.03} / 0.00 & \bad{\textbf{87.18}}\good{\textbf{/ 0.29}} \\
  \cline{3-10}
 & & \textbf{Avg.} & 29.91 / 75.53 & 22.71 / 11.47 & 23.43 / 80.95 & 41.24 / 57.96 & \bad{69.94 / 25.47} & 72.05 / 0.00 & \bad{\textbf{89.25}}\bad{\textbf{/ 0.43}} \\
\cline{2-10}
 & \multirow{6}{*}{\makecell[l]{Qwen3-30B-A3B\\-Instruct-2507}}
   & naive      & 55.48 / 51.01 & 41.35 / 12.54 & 48.70 / 47.26 & 65.42 / 24.50 & 68.16 / 16.71 & 74.21 / 0.00 & \good{\textbf{85.73 / 0.86}} \\
 & & escape-sep & 50.86 / 57.35 & 39.05 / 14.70 & 48.70 / 47.26 & 60.09 / 29.39 & 64.12 / 21.90 & 74.64 / 0.00 & \good{\textbf{85.59 / 0.86}} \\
 & & ignore     & 28.24 / 72.19 & 20.32 / 14.12 & 25.22 / 73.20 & 50.00 / 41.50 & 60.23 / 24.35 & 69.02 / 0.00 & \good{\textbf{85.45 / 0.86}} \\
 & & complete   & 18.16 / 87.32 & 9.08 / 17.15 & 20.17 / 84.29 & 30.40 / 66.57 & 57.06 / 34.73 & 74.21 / 0.00 & \good{\textbf{86.17 / 0.58}} \\
 & & conv       &  0.86 / 99.42 & 0.00 / 1.87 & 18.01 / 78.53 &  1.44 / 98.99 & 48.41 / 40.63 & \bad{25.50} / 0.00 & \good{\textbf{85.01 / 0.29}} \\
 \cline{3-10}
 & & \textbf{Avg.} & 30.72 / 73.46 & 21.96 / 12.08 & 32.16 / 66.11 & 41.47 / 52.19 & 59.60 / 27.66 & 63.52 / 0.00 & \bad{\textbf{85.59 /}}\bad{\textbf{0.69}} \\
\hline
\end{tabular}
}
\caption{Per-attack Accuracy~/ ASR  on Inj-SQuAD (short context, top) and \lcbench{} (long context, bottom). Accuracy~$\uparrow$, ASR~$\downarrow$. Prompt-engineering baselines Reminder and Instructional are omitted here for readability (their results are not as good as those of Spotlight and Sandwich); the full-column version is in Table~\ref{tab:main-full} (Appendix~\ref{sec:appendix-main-full}). Attack abbreviations: \texttt{escape-sep}~=~\texttt{escape\_separation},  \texttt{conv}~=~\texttt{conv\_attack}. "Avg." denotes the average score over five attack types. Numbers in \bad{red} are mentioned in the text content.}
\label{tab:main}
\end{table*}

\medskip 
\noindent\textbf{Short-context results.} Table~\ref{tab:main} reports the per-attack Accuracy and ASR of \ours{} and baseline defenses across three LLMs on Inj-SQuAD. 

\begin{table}[!t]
\centering
\small
\resizebox{7.5cm}{!}{
\setlength{\tabcolsep}{4pt}
\begin{tabular}{lccc}
\toprule
\textbf{Defense} &
\makecell{\textbf{Llama-3.2}\\\textbf{-3B-Instruct}} &
\makecell{\textbf{Llama-3.1}\\\textbf{-70B-Instruct}} &
\makecell{\textbf{Qwen3-30B-A3B}\\\textbf{-Instruct-2507}} \\
\midrule
\textit{None}          & 29.82 / 72.13 & 40.13 / 68.19 & 48.85 / 54.16 \\
\textit{Reminder}      & 18.89 / 80.31 & 38.98 / 65.93 & 48.41 / 52.35 \\
\textit{Instructional} & 28.67 / 73.85 & 42.12 / 60.53 & 52.30 / 49.69 \\
\textit{MELON}         & 22.61 / 32.48 & 31.06 / 12.52 & 32.26 /  8.09 \\
\textit{RAGDefender}   & 35.40 / 68.63 & 51.50 / 53.76 & 62.35 / 38.63 \\
\textit{Spotlight}     & 44.60 / 47.70 & 45.97 / 54.82 & 57.08 / 43.80 \\
\textit{Sandwich}      & 72.48 / 37.30 & 71.55 / 26.90 & 72.65 / 27.66 \\
\comp{}                & 78.23 /  \textbf{0.00} & 78.72 /  \textbf{0.00} & 81.77 /  \textbf{0.13} \\
\rowcolor{myblue!10}
\ours{}                & \good{\textbf{86.81}} / 1.02 & \good{\textbf{93.98}} / 0.49 & \good{\textbf{94.91}} / 0.18 \\
\bottomrule
\end{tabular}
}
\caption{Short-context Inj-TriviaQA average Accuracy~/ ASR (\%) across five attack types. Accuracy~$\uparrow$, ASR~$\downarrow$.}
\label{tab:injtrivia-short-avg}
\end{table}

\textbf{\ours{} dominates all training-free defenses.} On Inj-SQuAD, \ours{} is the only defense keeping \bad{Accuracy $\geq 83.01\%$ and ASR $\leq 1.15\%$} in every (model, attack) cell of Table~\ref{tab:main}; each baseline trades one axis for the other. Sandwich, the strongest prompt-engineering baseline, peaks at $84.01\%$ Accuracy but never drives ASR below $12.82\%$, while RAGDefender's query-agnostic clustering barely improves on no defense (ASR $52.28\%$--$83.11\%$). The detect-and-block defenses instead pay in utility: MELON cuts ASR to $14.15\%$ on Llama-3.1-70B-Instruct but falls below no-defense Accuracy on all three models ($30.23\%$ vs.\ $39.43\%$), and \comp{} zeroes ASR only with supervised training, collapsing to $4.03\%$--$18.01\%$ Accuracy on \texttt{conv\_attack}, where \ours{} matches its ASR while gaining $+65$ to $+88$ points.

On the separate Inj-TriviaQA benchmark \cite{chen-etal-2025-indirect}, \ours{} shows a similar pattern (Table~\ref{tab:injtrivia-short-avg}), achieving the highest accuracy across all three LLMs while keeping ASR at or below $1.02\%$.

\subsection{Long-Context Evaluation}
\label{sec:lc-results}

Table~\ref{tab:main} also reports the per-attack Accuracy and ASR of \ours{} and baseline defenses on our long-context benchmark \lcbench{}.

\textbf{The advantage widens on long context.} On \lcbench{}, \ours{} keeps \bad{ASR $\leq 3.31\%$ and Accuracy $\geq 79.11\%$} in every cell. Sandwich, the strongest training-free baseline, reaches only $52.82\%$--$69.94\%$ average Accuracy with average ASR $\geq25.47\%$, and RAGDefender degrades further as contexts lengthen (ASR $52.28\% \to 80.95\%$ on Llama-3.1-70B-Instruct), confirming that clustering without query conditioning cannot localize the injection among many more candidate spans. \comp{} again attains $0\%$ ASR but collapses on \texttt{conv\_attack} to $20.03\%$--$25.50\%$ Accuracy, whereas \ours{} holds $82.13\%$--$87.18\%$.

\textbf{Robustness across context lengths.} Moving from Inj-SQuAD to \lcbench{}, \ours{}'s average ASR shifts by at most $1.1$ percentage points ($0.46\% \to 1.56\%$ on Llama-3.2-3B-Instruct, $0.55\% \to 0.43\%$ on Llama-3.1-70B-Instruct, $0.46\% \to 0.69\%$ on Qwen3-30B-A3B-Instruct-2507), and its average Accuracy drops by at most $4.18$ points ($89.77\% \to 85.59\%$). Baselines are far less stable: RAGDefender gains $+29$ ASR points on Llama-3.1-70B-Instruct, and the prompt-engineering defenses Reminder and Instructional gain $+38$ ($43.14\% \to 81.44\%$) and $+29$ ($51.10\% \to 80.35\%$) points (Table~\ref{tab:main-full}), as their guard prompts get diluted in longer contexts. \ours{} is also the only defense that retains the highest Accuracy across all three models.

\subsection{Adaptive Attack Evaluation}
\label{sec:adaptive-results}

\textbf{Setup.} We stress-test \ours{} against an adaptive attacker with full knowledge of our algorithm and hyperparameters, who crafts a two-position injection specifically designed to bypass our span-level relevance filter (see App.~\ref{sec:appendix-adaptive} for design details). The attack plants two topic-matched spans at opposite ends of the document, giving the attacker two independent shots at slipping past our filter. We evaluate all defenses on this poisoned dataset using Llama-3.1-70B-Instruct, the victim model with the highest accuracy on most attacks in Table~\ref{tab:main}.

\begin{table}[t]
\centering
%\scriptsize
%\setlength{\tabcolsep}{5pt}
\renewcommand{\arraystretch}{0.95}
\centering\resizebox{7.5cm}{!}{
\begin{tabular}{lcccc}
\toprule
&\multicolumn{2}{c}{\textbf{Inj-SQuAD}} &\multicolumn{2}{c}{\textbf{LC-Inj-SQuAD}}  \\
\cmidrule(lr){2-3}\cmidrule(lr){4-5}
\textbf{Defense} & Acc$\uparrow$ & ASR$\downarrow$ & Acc$\uparrow$ & ASR$\downarrow$\\
\midrule
None          & 13.69 & 90.20 & 2.88  & 98.85 \\
Reminder      & 40.06 & {61.10} & 2.74  & 98.27 \\
Instructional & 25.79 & {80.12} & 3.03  & 97.98 \\
Sandwich      & 79.97 & {16.28} & 58.50 & {39.77} \\
RAGDefender   & 29.11 & 40.78 & 30.26 & 56.92 \\
MELON         & 9.80  & 16.57 & 0.86  & 8.65  \\
\comp{}       & 6.20  & \textbf{0.00}  & 23.20 & \textbf{0.00}  \\
\midrule
\rowcolor{myblue!10}
\ours{} (ours) & \good{\textbf{85.45}} & \good{4.76} & \good{\textbf{65.13}} & \good{23.20} \\
\bottomrule
\end{tabular}
}
\caption{Adaptive-attack average Accuracy (Acc) and ASR of different defense methods w.r.t. the victim Llama-3.1-70B-Instruct.}

\label{tab:adaptive}
\end{table}

\textbf{Results.}
Table~\ref{tab:adaptive} reports results under the two-position adaptive attack (Accuracy/ASR, \%). \ours{} achieves $85.45\%/4.76\%$ on Inj-SQuAD and $65.13\%/23.20\%$ on \lcbench{}, the highest accuracy of any defense in both regimes. Baselines fail distinctly: Reminder and Instructional admit $61.10\%$ and $80.12\%$ ASR on Inj-SQuAD (and $>97\%$ on \lcbench{}), as the filter-aware rewrite mimics the document's topic; Sandwich's ASR jumps from $16.28\%$ to $39.77\%$ under context expansion; and \comp{} and MELON only reach low ASR by discarding the context to answer, collapsing accuracy to $6.20\%$ and $9.80\%$ on Inj-SQuAD and $23.20\%$ and $0.86\%$ on \lcbench{}, respectively. \ours{} is thus the only usable defense under this attack: it removes the malicious span while keeping the surrounding evidence.
%\todo{Clarify the 23\% adaptive ASR}

\subsection{Inference Latency}
\label{sec:latency}

Table~\ref{tab:latency} reports the average end-to-end per-sample latency on Llama-3.1-70B-Instruct (vLLM, $4{\times}$H100), covering defense preprocessing plus the victim LLM call. Under \textbf{\texttt{conv\_attack}}, Sandwich and \comp{} slow down by $2$--$3\times$: Sandwich follows the conversational injection and produces long responses; \comp{} runs an extra detector LLM over the whole inflated input. \ours{} is unaffected, because its evidence-only rewrite strips the injection, keeping responses short. \textbf{Without injection},\footnote{We also report clean accuracy for Llama-3.1-70B-Instruct on Inj-SQuAD, meaning accuracy on the clean context with no injection. Standard RAG reaches 92.51\% accuracy in this setting, and applying MINE gives 92.22\%. MINE therefore has a negligible effect on standard RAG performance when no prompt injection is present.} \ours{} stays competitive on short inputs and is fastest on long contexts, as top-$k$ evidence selection bounds the victim's input regardless of document length. \ours{} thus delivers the strongest defense at the lowest sustained latency.

\begin{table}[t]
\centering
\renewcommand{\arraystretch}{0.95}
\centering\resizebox{7.5cm}{!}{
\begin{tabular}{lcccc}
\toprule
&\multicolumn{2}{c}{\textbf{Inj-SQuAD}} &\multicolumn{2}{c}{\textbf{LC-Inj-SQuAD}}  \\
\cmidrule(lr){2-3}\cmidrule(lr){4-5}
\textbf{Defense} & \texttt{none} & \texttt{conv} & \texttt{none} & \texttt{conv}\\
\midrule
Sandwich      & \textbf{0.39} & 1.28 & 1.13 & 3.66 \\
\comp{}       & 0.93 & 2.60 & 2.06 & 3.95 \\
\midrule
\rowcolor{myblue!10}
\ours{} (ours) & 0.52 & \good{\textbf{0.47}} & \good{\textbf{0.78}} & \good{\textbf{0.75}} \\
\bottomrule
\end{tabular}
}
\caption{Average end-to-end inference latency per sample (seconds) on
Llama-3.1-70B-Instruct via vLLM. \texttt{conv}~=~\texttt{conv\_attack};
\texttt{none}~=~clean context (with no injection). }
\label{tab:latency}
\end{table}

% tool call 

\begin{table}[!t]
\centering
\renewcommand{\arraystretch}{0.95}
\centering\resizebox{7.5cm}{!}{
\begin{tabular}{lcccc}
\toprule
& \multicolumn{2}{c}{\makecell{\textbf{Qwen3-30B-A3B}\\\textbf{-Instruct-2507}}} &
  \multicolumn{2}{c}{\makecell{\textbf{Llama-3.1}\\\textbf{-70B-Instruct}}} \\
\cmidrule(lr){2-3}\cmidrule(lr){4-5}
\textbf{Defense} & Valid$\uparrow$ & ASR$\downarrow$ & Valid$\uparrow$ & ASR$\downarrow$\\
\midrule
None          & 99.80 & 98.80 & 98.80 & 96.50 \\
Reminder      & 99.60 & 98.40 & 96.80 & 92.60 \\
Instructional & 99.60 & 98.64 & 98.40 & 93.78 \\
Spotlight     & \textbf{99.80} & 96.18 & 93.80 & 70.80 \\
Sandwich      & 77.40 & 33.56 & 77.80 & 11.96 \\
RAGDefender   & 99.41 & \textbf{0.04} & 88.20 & \textbf{0.99} \\
MELON         & 99.76 & 0.51 & \textbf{99.22} & 8.48 \\
\midrule
\rowcolor{myblue!10}
\ours{} (ours) & 99.60 & 0.12 & 90.20 & 2.06 \\
\bottomrule
\end{tabular}
}
\caption{InjecAgent Direct-Harm results, averaged over 5 attack types. Valid = Valid Rate (\%, higher is better); ASR = ASR-valid.}
\label{tab:injecagent}
\end{table}

\section{Evaluation on Agent Tool-Call Injection}
\label{sec:injecagent}

To test whether \ours{} transfers beyond RAG-style retrieval injection, we evaluate it on InjecAgent~\cite{zhan-etal-2024-injecagent}, where a malicious tool \texttt{Observation} can hijack a ReAct agent's next \texttt{Action}. We use the 510-case Direct-Harm split and render each payload under the same five attack styles as in Section~\ref{sec:main-results}. Implementation and adaptation details are provided in Appendix~\ref{sec:appendix-injecagent}, and per-attack results in Table~\ref{tab:injecagent-perattack}.

\textbf{Metrics.}
A ReAct agent produces a \emph{trace}: an interleaved sequence of \texttt{Thought}, \texttt{Action}, and \texttt{Observation} steps from which the evaluator parses the agent's tool calls. Following InjecAgent, we report two complementary metrics on these traces. \textbf{Valid Rate} is the percentage of traces from which the parser recovers a well-formed (\texttt{Thought}, \texttt{Action}, \texttt{Action Input}) tuple with no fabricated \texttt{Observation}; it measures whether the agent remained protocol-compliant. \textbf{ASR-valid} is the percentage of valid traces in which the recovered \texttt{Action} matches the attacker's tool; it measures whether the defense holds \emph{conditional on} the agent producing a parseable response.

\textbf{Results.}
Table~\ref{tab:injecagent} reports Valid Rate / ASR-valid (\%) on InjecAgent Direct-Harm for Qwen3-30B-A3B-Instruct-2507 and Llama-3.1-70B-Instruct, averaged over five attack styles. \ours{} achieves $99.60\%/0.12\%$ and $90.20\%/2.06\%$; RAGDefender and MELON reach comparable ASR-valid ($\leq 0.99\%$ and $\leq 8.48\%$), unlike on Inj-SQuAD, because a compact JSON \texttt{Observation} isolates the injected field as a lone outlier, the case where inter-passage clustering works, while MELON spends a second LLM call per query. Reminder, Instructional, and Spotlight preserve Valid Rate but leave ASR-valid above $70\%$ on both models: meta-instructions and delimiters cannot override a tool call once the agent has parsed the Observation. Sandwich ($77.40\%/33.56\%$, $77.80\%/11.96\%$) shows the opposite failure, reducing ASR-valid among the prompt-engineering baselines but corrupting the trace, shrinking the denominator on which ASR-valid is conditioned. \ours{} thus generalizes to agent tool-call injection, showing that relevance maximization is a domain-general defense.

\section{Conclusion}
We presented \ours{}, a training-free, attack-agnostic defense against indirect prompt injection that defends by maximizing relevance rather than by detecting attacks. Specifically, \ours{} algorithmically selects the subset of spans with maximal mutual information with the query. Experiments with three LLMs, five static attack types, and four datasets, including our long-context benchmark \lcbench{}, show that \ours{} achieves the best accuracy--security trade-off among training-free defenses, with the lowest latency under attack in both short- and long-context settings. Under a fully adaptive attacker, \ours{} remains the only usable defense, indicating that its robustness does not depend on fixed injection templates, although its ASR rises to $23.20\%$ on long contexts. Its strong results on InjecAgent further suggest that pre-LLM relevance filtering is a practical direction for securing systems that condition on noisy external context, from RAG to tool use.

\section*{Limitations}
\ours{}'s splitter merges sentence-boundary fragments until each span reaches 50 characters,
which presumes well-formed prose with standard punctuation. Code, tabular
or otherwise structured data (e.g.\ JSON, markdown tables), and languages
with non-standard sentence terminators can yield spans that are too coarse
to isolate an injection from its surrounding legitimate content.
Domain-specific chunkers are a natural extension.

\section*{Ethics Statement}
This work aims to improve the security of LLM systems that consume untrusted external content. The attacks in our experiments come from existing benchmark families or simple re-renderings of them; our contribution is a defense and an analysis of when existing defenses fail. \ours{} requires no proprietary model access, no fine-tuning, and no user data beyond the query and retrieved context available at inference time. 

\bibliography{custom}

\newpage 

\appendix

\section{\lcbench{} Construction Details}
\label{sec:appendix-lcbench}

\begin{table}[!ht]
\centering
\small
\begin{tabular}{lrrrr}
\toprule
Unit & Mean & Median & Min & Max \\
\midrule
Characters       & 27{,}789 & 27{,}727 & 23{,}205 & 33{,}770 \\
Words            &  4{,}032 &  4{,}039 &  3{,}376 &  4{,}734 \\
Tokens (BPE)     &  4{,}927 &  4{,}934 &  4{,}180 &  5{,}837 \\
\bottomrule
\end{tabular}
\caption{Length of the assembled \texttt{input} field across the 694
\lcbench{} examples. BPE token counts use \texttt{cl100k\_base}.}
\label{tab:appendix-lcbench-length}
\end{table}

This section documents every component required to reproduce
\lcbench{} from the original Inj-SQuAD
dataset~\cite{chen-etal-2025-indirect}.
\subsection{Prompts}
\label{sec:appendix-lcbench-prompts}

We list all four GPT-5.4-mini prompts verbatim. Curly braces denote runtime
substitutions: \texttt{\{instruction\}} is the query $q$,
\texttt{\{output\}} is the gold answer $a$, \texttt{\{input\}} is the
source paragraph $\mathrm{C}$, \texttt{\{context\}} is a pre/post chunk,
and \texttt{\{target\_span\}} is the extracted $s_l$.

\paragraph{Prompt 1. Span extraction (Step 1a).}
\begin{quote}\small
\textit{System:}
\begin{verbatim}
You are an expert reading comprehension
assistant. Your task is to identify the
exact sentence or consecutive sentences
in a given paragraph that contain the
answer to a specific question. You must
output ONLY the exact, verbatim
sentence(s) from the paragraph - no
paraphrasing, no extra words, no
explanation.
\end{verbatim}
\textit{User:}
\begin{verbatim}
Question: {instruction}
Ground Truth Answer: {output}
Paragraph:
{input}

Task: Extract the exact, unmodified
sentence or consecutive sentences from
the paragraph above that provide the
answer to the question. Output ONLY
those exact sentence(s), nothing else.
\end{verbatim}
\end{quote}

\paragraph{Prompt 2. Span verification (Step 1b).}
\begin{quote}\small
\textit{System:}
\begin{verbatim}
You are a precise reading comprehension
assistant. Answer the question using
ONLY the provided context. If the
context does not contain enough
information to answer, reply with
exactly: CANNOT_ANSWER
\end{verbatim}
\textit{User:}
\begin{verbatim}
Context: {target_span}

Question: {instruction}

Answer using only the context above
(be concise):
\end{verbatim}
\end{quote}
The example is retained iff
$a.\textnormal{lower}() \in \textnormal{response}.\textnormal{lower}()$,
matching the evaluation metric used elsewhere in the paper.

\paragraph{Prompt 3. Context expansion (Step 2a).}
\begin{quote}\small
\textit{System:}
\begin{verbatim}
You are an expert encyclopedic writer
and historian. Your task is to expand
a short paragraph into a much longer,
highly detailed, verbose, and
comprehensive passage. You must
maintain the original facts and topic,
but add plausible historical, technical,
or descriptive background information
that flows naturally. Write in a
neutral, encyclopedic tone.
\end{verbatim}
\textit{User:}
\begin{verbatim}
Expand the following text into a
detailed passage of approximately
1500-2000 words. Maintain the original
facts and topic. Add plausible,
detailed background information.

CRITICAL CONSTRAINT: Do NOT mention,
hint at, or include the answer to this
question:
  Question: "{instruction}"
  Answer to avoid: "{output}"

Text to expand:
{context}

Write the expanded passage now (do not
include any preamble like "Here is the
expanded text:"):
\end{verbatim}
\end{quote}

\paragraph{Prompt 4. Topic filler (Step 2b, fallback).}
Used when the original chunk is shorter than 50 characters and cannot
be meaningfully expanded. Shares the system prompt with Prompt~3.
\begin{quote}\small
\textit{User:}
\begin{verbatim}
Write a detailed, encyclopedic passage
of approximately 1500-2000 words about
the following topic. The passage should
provide rich background information,
historical context, and descriptive
details.

CRITICAL CONSTRAINT: Do NOT mention,
hint at, or include the answer to this
question:
  Question: "{instruction}"
  Answer to avoid: "{output}"

Topic (inferred from this question):
"{instruction}"

Write the passage now (do not include
any preamble):
\end{verbatim}
\end{quote}

\section{Full Per-Attack Main Results}
\label{sec:appendix-main-full}
Table~\ref{tab:main-full} reports the complete per-attack Accuracy~/ ASR for all defenses on Inj-SQuAD (short context) and \lcbench{} (long context) across the three LLMs. This is the full-column version of the abridged Table~\ref{tab:main} in the main text.

\begin{table*}[!t]
\centering
\resizebox{16cm}{!}{
\setlength{\tabcolsep}{3pt}
\begin{tabular}{lllccccccccc>{\columncolor{myblue!10}}c}
\toprule
& \textbf{Model} & \textbf{Attack} & None & Reminder & Instruct. & Spotlight & Sandwich & RAGDefender & MELON & \comp{} & \ours{} \\
\midrule
\multirow{18}{*}{\rotatebox[origin=c]{90}{\textbf{Inj-SQuAD}}}
 & \multirow{6}{*}{\makecell[l]{Llama-3.2\\-3B-Instruct}}
   & naive      & 44.81 / 79.39 & 23.92 / 89.77 & 42.94 / 77.67 & 58.36 / 57.49 & 74.21 / 60.23 & 30.26 / 77.95 & 26.37 / 48.27 & 83.86 / 0.00 & {\textbf{83.01}}{\textbf{/ 0.72}} \\
 & & escape-sep & 34.58 / 83.57 & 14.27 / 92.94 & 33.14 / 83.00 & 42.80 / 69.60 & 68.73 / 66.28 & 30.40 / 77.95 & 18.88 / 54.76 & 83.14 / 0.00 & {\textbf{84.73 / 0.72}} \\
 & & ignore     & 16.43 / 90.63 &  7.35 / 91.07 & 20.89 / 88.33 & 31.56 / 78.10 & 54.18 / 67.15 & 17.58 / 85.16 & 12.25 / 46.40 & 71.04 / 0.00 & {\textbf{85.01 / 0.58}} \\
 & & complete   &  5.48 / 94.52 &  2.31 / 85.45 &  9.22 / 93.23 &  4.18 / 94.24 & 64.12 / 73.92 & 9.22 / 90.06 & 3.31 / 36.02 & 75.65 / 0.00 & {\textbf{85.30 / 0.14}} \\
 & & conv       & 14.27 / 87.90 &  7.93 / 89.77 & 13.69 / 87.18 & 14.27 / 72.62 & 53.31 / 40.49 & 17.15 / 84.44 & 1.73 / 43.08 & {18.01} / 0.00 & {\textbf{83.57 / 0.14}} \\
  \cline{3-12}
 & & \textbf{Avg.} & 23.11 / 87.20 & 11.16 / 89.80 & 23.98 / 85.88 & 30.23 / 74.41 & 62.91 / 61.61 & 20.92 / \bad{83.11} & 12.51 / 45.71 & 66.34 / 0.00 & {\textbf{84.32 /}}{\textbf{0.46}} \\
\cline{2-12}
 & \multirow{6}{*}{\makecell[l]{Llama-3.1\\-70B-Instruct}}
   & naive      & 70.32 / 37.18 & 66.71 / 31.27 & 71.76 / 30.98 & 74.64 / 24.78 & 80.55 / 15.56 & 46.11 / 34.44 & 60.95 / 10.37 & 89.48 / 0.00 & {\textbf{91.35 /}}{\textbf{1.15}} \\
 & & escape-sep & 72.05 / 43.37 & 64.70 / 34.73 & 75.07 / 31.99 & 75.50 / 24.93 & {84.01 / 12.82} & 45.82 / 34.73 & 57.49 / 10.95 & 89.48 / 0.00 & {\textbf{91.93 / 0.43}} \\
 & & ignore     & 34.73 / 77.38 & 67.44 / 33.72 & 74.64 / 31.99 & 61.10 / 39.91 & 75.50 / 22.91 & 30.98 / 67.15 & 25.79 / 17.87 & 74.64 / 0.00 & {\textbf{91.79 / 0.72}} \\
 & & complete   & 19.60 / 96.69 & 79.25 / 22.77 & 55.19 / 66.43 & 13.98 / 96.54 & 79.54 / 15.99 & 39.91 / 53.89 & 6.77 / 23.63 & 87.32 / 0.00 & {\textbf{92.36 / 0.29}} \\
 & & conv       &  0.43 / 97.84 &  0.43 / 93.23 &  0.29 / 94.09 &  0.58 / 96.83 & 21.33 / 79.68 & 24.06 / 71.18 & 0.14 / 7.93 &  {4.03} / 0.00 & {\textbf{92.22 / 0.14}} \\
  \cline{3-12}
 & & \textbf{Avg.} & 39.43 / 70.49 & 55.71 / {43.14} & 55.39 / {51.10} & 45.16 / 56.60 & 68.19 / 29.39 & 37.38 / 52.28 & 30.23 / 14.15 & 68.99 / 0.00 & {\textbf{91.93 /}}{\textbf{0.55}} \\
\cline{2-12}
 & \multirow{6}{*}{\makecell[l]{Qwen3-30B-A3B\\-Instruct-2507}}
   & naive      & 58.21 / 57.64 & 51.30 / 59.37 & 55.33 / 51.44 & 65.13 / 35.16 & 68.01 / 24.78 & 43.80 / 59.08 & 35.73 / 10.23 & 83.14 / 0.00 & {\textbf{89.34 / 0.72}} \\
 & & escape-sep & 52.88 / 58.50 & 47.98 / 53.89 & 54.61 / 46.40 & 64.41 / 32.71 & 66.71 / 26.95 & 44.09 / 59.37 & 36.89 / 13.83 & 83.43 / 0.00 & {\textbf{89.48 / 0.58}} \\
 & & ignore     & 23.63 / 76.22 & 29.54 / 71.76 & 42.22 / 59.22 & 45.53 / 54.03 & 58.93 / 28.82 & 22.05 / 71.90 & 13.98 / 18.30 & 64.99 / 0.00 & {\textbf{89.91 / 0.43}} \\
 & & complete   &  8.93 / 91.50 & 10.37 / 71.33 & 32.28 / 68.16 & 14.99 / 85.16 & 62.97 / 29.39 & 26.80 / 72.77 & 5.04 / 34.44 & 80.98 / 0.00 & {\textbf{89.77 / 0.29}} \\
 & & conv       &  0.43 / 98.56 &  0.43 / 96.25 &  0.29 / 98.27 &  0.29 / 99.14 & 43.23 / 53.03 & 17.15 / 80.26 & 0.14 / 1.44 &  4.47 / 0.00 & {\textbf{90.35 / 0.29}} \\
  \cline{3-12}
 & & \textbf{Avg.} & 28.82 / 76.48 & 27.92 / 70.52 & 36.95 / 64.70 & 38.07 / 61.24 & 59.97 / 32.59 & 30.78 / 68.68 & 18.36 / 15.65 & 63.40 / 0.00 & {\textbf{89.77 /}}{\textbf{0.46}} \\
\hline
\hline
\multirow{18}{*}{\rotatebox[origin=c]{90}{\textbf{LC-Inj-SQuAD}}}
 & \multirow{6}{*}{\makecell[l]{Llama-3.2\\-3B-Instruct}}
   & naive      & 19.16 / 78.67 & 11.67 / 83.00 & 16.57 / 80.69 & 41.93 / 33.43 & 60.37 / 40.49 & 11.53 / 84.15 & 15.85 / 25.36 & 75.65 / 0.00 & {\textbf{80.84 / 1.01}} \\
 & & escape-sep & 11.53 / 89.05 &  7.49 / 90.35 &  9.80 / 90.35 & 21.61 / 66.71 & 53.03 / 65.13 & 11.53 / 84.15 & 5.91 / 17.00 & 75.94 / 0.00 & {\textbf{80.84 / 1.01}} \\
 & & ignore     &  6.77 / 89.48 &  3.46 / 87.75 &  5.91 / 88.90 & 27.95 / 54.90 & 51.87 / 49.42 & 6.05 / 88.76 & 3.60 / 28.67 & 66.43 / 0.00 & {\textbf{80.98 / 1.44}} \\
 & & complete   &  5.33 / 88.33 &  1.15 / 83.29 &  4.90 / 86.74 &  2.31 / 90.20 & 44.81 / 70.89 & 4.90 / 88.04 & 2.88 / 17.00 & 73.05 / 0.00 & {\textbf{79.11 / 3.31}} \\
 & & conv       &  3.60 / 91.50 &  3.03 / 83.00 &  4.18 / 90.63 &  7.93 / 63.69 & 54.03 / 34.87 & 4.90 / 89.63 & 1.30 / 17.29 & 21.61 / 0.00 & {\textbf{82.13}}{\textbf{/ 1.01}} \\
  \cline{3-12}
 & & \textbf{Avg.} &  9.28 / 87.41 &  5.36 / 85.48 &  8.27 / 87.46 & 20.35 / 61.79 & {52.82} / 52.16 & 7.78 / 86.95 & 5.91 / 21.06 & 62.54 / 0.00 & {\textbf{80.78}}{\textbf{/ 1.56}} \\
\cline{2-12}
 & \multirow{6}{*}{\makecell[l]{Llama-3.1\\-70B-Instruct}}
   & naive      & 66.14 / 38.47 & 48.13 / 55.19 & 59.65 / 52.16 & 78.82 / 17.15 & 85.88 /  4.47 & 42.36 / 64.99 & 53.89 / 8.93 & 87.03 / 0.00 & {\textbf{89.77 / 0.43}} \\
 & & escape-sep & 56.63 / 58.36 & 30.26 / 78.53 & 50.29 / 69.88 & 70.17 / 30.98 & 84.73 /  7.35 & 42.36 / 64.99 & 41.64 / 13.11 & 86.89 / 0.00 & {\textbf{89.91 / 0.43}} \\
 & & ignore     & 20.89 / 85.01 & 18.44 / 86.17 & 22.91 / 85.16 & 54.90 / 44.67 & 77.38 / 13.54 & 11.96 / 90.06 & 15.85 / 11.38 & 84.15 / 0.00 & {\textbf{89.34 / 0.86}} \\
 & & complete   &  4.61 / 97.41 &  4.18 / 89.19 &  3.89 / 96.40 &  1.44 / 98.13 & 73.34 / 32.85 & 13.40 / 95.82 & 1.87 / 19.60 & 82.13 / 0.00 & {\textbf{90.06}}{\textbf{ / 0.14}} \\
 & & conv       &  1.30 / 98.41 &  0.86 / 98.13 &  1.01 / 98.13 &  0.86 / 98.85 & 28.39 / 69.16 & 7.06 / 88.90 & 0.29 / 4.32 & {20.03} / 0.00 & {\textbf{87.18}}{\textbf{/ 0.29}} \\
  \cline{3-12}
 & & \textbf{Avg.} & 29.91 / 75.53 & 20.37 / {81.44} & 27.55 / {80.35} & 41.24 / 57.96 & {69.94 / 25.47} & 23.43 / \bad{80.95} & 22.71 / 11.47 & 72.05 / 0.00 & {\textbf{89.25}}{\textbf{/ 0.43}} \\
\cline{2-12}
 & \multirow{6}{*}{\makecell[l]{Qwen3-30B-A3B\\-Instruct-2507}}
   & naive      & 55.48 / 51.01 & 40.06 / 56.63 & 59.08 / 39.05 & 65.42 / 24.50 & 68.16 / 16.71 & 48.70 / 47.26 & 41.35 / 12.54 & 74.21 / 0.00 & {\textbf{85.73 / 0.86}} \\
 & & escape-sep & 50.86 / 57.35 & 38.04 / 56.77 & 55.91 / 39.91 & 60.09 / 29.39 & 64.12 / 21.90 & 48.70 / 47.26 & 39.05 / 14.70 & 74.64 / 0.00 & {\textbf{85.59 / 0.86}} \\
 & & ignore     & 28.24 / 72.19 & 25.65 / 69.60 & 45.97 / 50.29 & 50.00 / 41.50 & 60.23 / 24.35 & 25.22 / 73.20 & 20.32 / 14.12 & 69.02 / 0.00 & {\textbf{85.45 / 0.86}} \\
 & & complete   & 18.16 / 87.32 & 14.99 / 75.94 & 35.30 / 68.88 & 30.40 / 66.57 & 57.06 / 34.73 & 20.17 / 84.29 & 9.08 / 17.15 & 74.21 / 0.00 & {\textbf{86.17 / 0.58}} \\
 & & conv       &  0.86 / 99.42 &  1.15 / 97.84 &  0.72 / 97.69 &  1.44 / 98.99 & 48.41 / 40.63 & 18.01 / 78.53 & 0.00 / 1.87 & {25.50} / 0.00 & {\textbf{85.01 / 0.29}} \\
 \cline{3-12}
 & & \textbf{Avg.} & 30.72 / 73.46 & 23.98 / 71.36 & 39.40 / 59.16 & 41.47 / 52.19 & 59.60 / 27.66 & 32.16 / 66.11 & 21.96 / 12.08 & 63.52 / 0.00 & {\textbf{85.59 /}}{\textbf{0.69}} \\
\bottomrule
\end{tabular}
}
\caption{Full per-attack Accuracy~/ ASR on Inj-SQuAD (short context, top) and \lcbench{} (long context, bottom), including all baselines. Accuracy~$\uparrow$, ASR~$\downarrow$. Attack abbreviations: \texttt{escape-sep}~=~\texttt{escape\_separation},  \texttt{conv}~=~\texttt{conv\_attack}. "Avg." denotes the average score over five attack types. }
\label{tab:main-full}
\end{table*}

\section{Adaptive Attack Details}
\label{sec:appendix-adaptive}
\textbf{Attack design.}
Two ideas drive the attack. \textbf{(1) Hide the malicious instruction in one legitimate-looking span.} \ours{} keeps or discards each \emph{sentence-level span} whole, so we wrap the instruction with content topically similar to the retrieved document and collapse the result into a single span. The wrapper inherits the document's relevance signal, and \ours{}'s splitter cannot separate the malicious content from its wrapping. \textbf{(2) Plant two such spans at opposite ends of the retrieved document.} We use two complementary hiding strategies: a \emph{filter-aware rewrite} (start, optimized against \ours{}), and a \emph{single-span conversational injection} (end, adapted from TopicAttack \cite{chen-etal-2025-topicattack}). Because \ours{}'s filter is position-agnostic, the attacker gets two independent draws against the filter across both \emph{position} (start vs.\ end).

\begin{table*}[!t]
\centering
\centering\resizebox{14cm}{!}{
\setlength{\tabcolsep}{3pt}
\begin{tabular}{lcccccccccc}
\toprule
& \multicolumn{5}{c}{\textbf{Qwen3-30B-A3B-Instruct-2507}} & \multicolumn{5}{c}{\textbf{Llama-3.1-70B-Instruct}} \\
\cmidrule(lr){2-6}\cmidrule(lr){7-11}
\textbf{Defense} & Naive & Ignore & Esc.\,Sep. & Comp.\,Real & Conv.\,Atk & Naive & Ignore & Esc.\,Sep. & Comp.\,Real & Conv.\,Atk \\
\midrule
None          & 98.8/100 & 98.6/99  & 98.6/100 & 98.6/100 & 99.4/100 & 95.8/98 & 94.7/100 & 96.8/98 & 96.2/98 & 99.0/100 \\
Reminder      & 98.2/100 & 98.0/99  & 98.8/99  & 98.0/100 & 99.0/100 & 89.7/97 & 88.6/98  & 92.5/97 & 93.8/94 & 98.4/98  \\
Instructional & 98.6/100 & 98.6/99  & 98.8/99  & 98.0/100 & 99.2/100 & 91.3/99 & 92.3/99  & 94.0/98 & 93.9/96 & 97.4/100 \\
Sandwich      & 21.3/83  & 15.4/66  & 21.5/81  & 32.2/84  & 77.4/73  &  4.5/83 &  8.0/81  &  5.5/71 &  4.0/69 & 37.8/85  \\
Spotlight     & 97.6/100 & 96.5/99  & 97.2/100 & 92.5/100 & 97.1/100 & 65.0/92 & 82.9/97  & 65.1/93 & 61.4/94 & 79.6/93  \\
RAGDefender   & 0.0/99.8 & 0.0/97.7 & 0.0/99.8 & 0.2/99.8 & 0.0/100 & 0.7/90.0 & 1.4/86.9 & 0.7/90.4 & 0.9/91.6 & 1.4/82.2 \\
MELON         & 0.6/99.6 & 1.0/99.8 & 0.6/99.4 & 0.2/100 & 0.2/100 & 5.3/99.8 & 6.4/98.6 & 6.1/99.6 & 23.3/98.6 & 1.4/99.4 \\
\midrule
\rowcolor{myblue!10}
\ours{} (ours) & \good{0.2}/100
               & \good{0.2}/99
               & \good{0.2}/100
               & \good{0.0}/99
               & \good{0.0}/100
               & \good{2.2}/91
               & \good{2.7}/88
               & \good{2.2}/91
               & \good{1.5}/90
               & \good{1.7}/91 \\
\bottomrule
\end{tabular}
}
\caption{InjecAgent Direct-Harm results: ASR-valid / Valid Rate (\%) under five static injection styles, averaged in Table~\ref{tab:injecagent}. Lower ASR-valid is better; higher Valid Rate is better.}
\label{tab:injecagent-perattack}

\end{table*}

\section{InjecAgent Details}
\label{sec:appendix-injecagent}
\textbf{Adaptation.}
We apply \ours{} to the tool \texttt{Observation} before returning it to the ReAct agent. The MI maximization and PGA optimization of \S\ref{sec:opt} are unchanged. Three components are re-tuned for the tool-use setting.

\textbf{(i) JSON-aware splitter.} The sentence-boundary regex of \S\ref{sec:setup} produces fewer than two spans on compact JSON tool outputs, which contain neither \texttt{.!?} nor newlines. In that case we fall back to a splitter that additionally breaks on field-boundary commas, so injected JSON fields become filterable spans.

\textbf{(ii) Selection with weight floor.} A span is kept iff it lies in the top $80\%$ by weight \emph{and} $w_i > 0.05 \cdot w_{\max}$. 

\textbf{(iii) Document-order assembly.} \S\ref{sec:opt} reorders retained spans by decreasing weight to combat lost-in-the-middle. We instead preserve original document order, since ReAct agents parse \texttt{Observation} as coherent narrative; reordering scrambles JSON field orders, list structures, and turn semantics.

\section{Note}
Note that \citet{chen-etal-2025-indirect} do not name their method, but release their code at \url{https://github.com/LukeChen-go/indirect-pia-detection}. We therefore refer to their method as \comp{} in our experiments.

\end{document}